\documentclass[conference]{IEEEtran}%{article}
\usepackage[utf8]{inputenc} % allow utf-8 input
\usepackage[T1]{fontenc}    % use 8-bit T1 fonts
\usepackage{hyperref}       % hyperlinks
\usepackage{url}            % simple URL typesetting
\usepackage{booktabs}       % professional-quality tables
\usepackage{amsfonts}       % blackboard math symbols
\usepackage{nicefrac}       % compact symbols for 1/2, etc.
\usepackage{graphicx}
\usepackage{amssymb}
\usepackage{cite}
\usepackage{hyperref}
\usepackage{soul, color}
\usepackage{caption}
\usepackage{subcaption}
\usepackage{float}
\graphicspath{ {./images/} }

\title{Distributed Conditional GAN (discGAN) for Synthetic Healthcare Data Generation}

\author{
 David Fuentes \\
  School of Data Science\\
  University of Virginia\\
  \texttt{dmf4ns@virginia.edu} \\
   \and
 Diana McSpadden \\
  School of Data Science\\
  University of Virginia\\
  \texttt{hdm5s@virginia.edu}
    \and
 Sodiq Adewole \\
  School of Data Science\\
  University of Virginia\\
  \texttt{soa2wg@virginia.edu} 
}

\begin{document}
\maketitle

\begin{abstract}
In this paper, we propose a distributed Generative Adversarial Networks (discGANs)  to generate synthetic tabular data specific to the healthcare domain. While using GANs to generate images has been well studied, little to no attention has been given to generation of tabular data. Modeling distributions of discrete and continuous tabular data is a non-trivial task with high utility. We applied discGAN to model non-Gaussian multi-modal healthcare data. We generated 249,000 synthetic records from original 2027 eICU dataset and we evaluated the performance of the model using machine learning efficacy, the Kolmogorov–Smirnov test (KS test) for continuous variables, and chi-squared test for discrete variables. Our results show that discGAN was able to generate data with distributions similar to the real data.
\end{abstract}
% keywords can be removed

%Our DS6050 Deep Learning project team coalesced 

% \section{GitHub Repository}
% Repository contains all files related to our DS6050 project on generating healthcare data using our distributed, conditional GAN (discGAN) as well as files associated with distributing CTGAN.

% \url{https://github.com/dianamJLAB/DS6050}

\section{Introduction}
%Our objective was to build generative adversarial networks (GANs) to create synthetic tabular %data specific to the healthcare domain using a distributed algorithm. While using GANs to %generate images is well studied, with 31 Papers with Code~\cite{pwc} results for `GAN images' %and over 1 million Google Scholar results, only nine Papers With Code results are available for %`tabular GAN', including no results specific to a distributed tabular GAN. We add to the GAN %body of knowledge by demonstrating the union of tabular and distributed GANs.

\label{section:motivation}
\subsection{Healthcare data: privacy concerns and utility}
Healthcare information falls into a unique category of protected health information under the Health Insurance Portability and Accountability Act (HIPPA). This poses unique challenges to healthcare research, healthcare administration research, and open science. Anonymous, publicly-available healthcare information exist, such as MIT's eICU Collaborative Research Database\cite{pollard2019eicu, eICU, physio}. However, only 2,500 intensive care unit stays are represented in the publicly available, demo version of the database; thus, the ability to generate statistically similar synthetic data is useful for open-science research. 

%\paragraph{Tabular healthcare data: useful for medical research} 
The complex dynamics between hospital stay characteristics -- such as pre-existing conditions, medications, and patient demographics -- may contribute to individual patient outcomes, and efficient and effective use of hospital resources. Using data science methods to understand these complex dynamics requires sufficient data to develop algorithms, and data science algorithms require adequate statistics across the parameter space. Tabular healthcare data includes continuous features, binary and multi-class discrete features, and could even include medical images. We address a tabular GAN for discrete and continuous variables. Generating statistically-similar hospital and patient data conditioned on hospital and medical characteristics has high utility for healthcare research.

\subsection{Synthetic healthcare data: a solution?}
The healthcare research community has been able to effectively close the door on public patient-scale research due to a lack of anonymous healthcare data. The contributions that could be made in hospital resource utilization, anomaly identification in patient care or hospital usage, and characterizing patient risk factors are significant, but only possible if enough data is available to feed data-hungry algorithms. We propose that GANs present a potential solution to these problems. With robust GAN architectures, researchers could leverage anonymous synthetic data for public health benefits without exposing the underlying individuals in the raw data. 

\label{section:litreview}
\section{Literature Survey}

%\subsection{Data privacy guarantees ~\cite{huang2019generative, brunel2020propose}} 

%Of primary interest is whether one can assume that generated data maintains privacy or, if %through reverse-engineering of single or multiple generated datasets, the private, real data can %be exposed. Both~\cite{huang2019generative} and ~\cite{brunel2020propose} address creating %output from a highly-sensitive input of $n$ observations such that the sensitive input data are %protected without a loss of the global characteristics surrounding those input data. %\cite{huang2019generative} proposes a method based on GANs, while \cite{brunel2020propose} %proposes a novel algorithm with a differential privacy guarantee for data without [known] %bounds, and without an assumption that mean or median estimators are within known bounds. While %\cite{huang2019generative} and \cite{brunel2020propose} are not within the scope of our project %as it currently stands, they were useful for understanding the mathematical analysis of privacy %since health data should be protected and would likely be considered protected health %information (PHI) if handled by an entity covered by the HIPPA~\cite{williams_tien}.

%Diana
\subsection{CTGAN: Modeling Conditional Tabular Data from Synthetic Data Vault} \cite{patki2016synthetic}

From MIT, CTGAN is the SOTA tabular data generator with the ability to generate discrete and continuous values, model non-Gaussian multi-model data, and learn conditional distributions for many dimensions. \cite{xu2019modeling} builds on previous tabular GAN implementations such as VeeGAN\cite{srivastava2017veegan}, TableGAN\cite{park2018data}, and MedGAN\cite{choi2017generating}. CTGAN's contributions include "augmenting the training procedure with mode-specific normalization" to address multi-modal continuous columns. For each continuous column a variational Gaussian mixture model estimates the number of modes, then for each record in the column the probability of the value coming from each mode is calculated, then the sampled mode from the probability density is used for normalization of the value. Ultimately, the column value is represented as a scalar and a one-hot vector indicating which mode the value is from.

Additionally, CTGAN adds a conditional generator and training-by-sampling to address a lack of samples from minor categories during training, i.e. unbalanced data. CTGAN samples in a way "that all the categories from discrete attributes are sampled evenly (though not necessarily uniformly)" during training. CTGAN learns a conditional vector indicating which condition should be sampled. During training the generator is penalized for generating data that does not match an indicated conditional vector. CTGAN's training-by-sampling approach ensures that the model "evenly explore[s] all possible values in discrete columns", even minor categories.%We appreciate the evaluation process of CTGAN. 
For benchmark datasets of real data, \emph{machine learning efficacy} was used to evaluate performance. A set of test data was held out during training, and classification and regression models were trained on generated data, then tested on the holdout data. Classification tasks used accuracy and F1 metrics, and regression tasks used $R^2$.

% Diana
\subsection{"Md-gan: Multi-discriminator generative adversarial networks for distributed datasets"}  
The earliest of the papers reviewed describing architectures for distributed GANs, the approach in\cite{hardy2019md} proposes a single, centralized generator, and multiple, distributed discriminator workers. Md-gan considered a scenario of $N$ workers, possibly geographically distributed across datacenters, each with their own local "truth" training dataset (assumed i.i.d.). Local datasets remain on their worker, which limits the bandwidth needed for Md-gan when compared to an architecture that would require dataset sharing. 

Md-gan evaluated its performance on communication and computation complexity, memory, and quality against a standalone GAN and FL-GAN -- a distributed GAN with both a generator and discriminator on each worker that had their model weights averaged throughout training. Md-gan was solely evaluated on generating images. Md-gan was competitive on communication complexity for smaller batch sizes, less than 550, but with non-image data, batch size may behave differently. The paper reports an expectation that Md-gan has half the computation complexity of FL-GAN due to a single generator vs. a generator per worker; the complexity is not 1/$N$ because the consequence of the single, central generator is more frequent interactions between the workers and the generator.

\cite{hardy2019md} evaluated performance generating images from the MNIST, CIFAR10, and CelebA datasets using Inception Score (IS), and Fréchet Inception Distance (FID). \cite{hardy2019md} describes IS as an evaluation of the generated data when classified on a pre-trained Inception classifier. IS evaluates whether the generated data are "well recognized" and diverse. A higher IS score is better. FID measures a "distance between the distribution of the generated data and real data." To evaluate this distance, an Inception network or a network adapted from an Inception network is used, making this metric image specific. 
%A lower FID score is better. 

%Comparing the three image datasets, and against both a standalone GAN and %FL-GAN, Md-gan performed well with the lowest FID on CIFAR10 and MNIST MLP %datasets and second lowest FID on MNIST-fashion. Using an MLP on MNIST and %a CNN on CIFAR10, Md-gan performed better for both IS and FID metrics, and %for IS on the CelebA dataset. The standalone GAN performed best on a CNN %for MNIST for both metrics, and comparing FID on the CelebA dataset. The %FL-GAN did not perform best using any metric on any dataset.

Additionally, Md-gan evaluated GAN performance with and without peer-to-peer discriminator swapping throughout training. There was only a marginal gain in performance when periodically swapping discriminator weights between workers. Fault tolerance to worker crashes was also described. The authors report "the MD-GAN architecture manages to learn fast enough so that crashes, and then the removal of dataset shares, are not a problem performance wise."

% Diana
\subsection{"FeGAN: Scaling Distributed GANs"}

\cite{fegan} demonstrates a distributed, conditional GAN with distribution of both the generator (G) and discriminator (D) on K devices. Different devices have access to different samples from different subsets of the total classes. Devices are asked to contribute based on their capabilities, i.e. memory and compute power. Devices must communicate to the server their capability and the number of samples per class. The central server will "load control" requests to devices to balance the contribution of device G's and D's trained on different class distributions with differing sample counts.

A central server starts each round of learning by selecting a group of devices from K. Devices are selected to "ensure a balanced number of samples per class". Only a subset of the total devices may be used during a training round. The server also provides selected devices with the current model state. The selected devices train for a configured number of epochs, then send updated models to the server. After receiving updated G's and D's the server aggregates the models using a "KL weighting scheme" by "applying the softmax function on the KL divergence score of each device." This scheme promotes device model updates with data distributions similar to the global data distribution, and penalizes device model updates with greater divergence from the global data distribution. This method attempts to avoid mode collapse.

\cite{fegan} evaluated performance generating images from the MNIST, Fashion-MNIST, and ImageNet using Fréchet Inception Distance (FID). \cite{fegan} reports that the co-location of G and D on devices, KL weighting for model aggregation, and balanced sampling result in a generator powerful enough to generate multimodal distributions, demonstrated by their lower FID values compared to a vanilla federated learning GAN. 
%Additionally, \cite{fegan} evaluated time to convergence, scalability of distributed devices, and bandwidth consumption, i.e. throughput, compared %favorably to Md-gan~\cite{hardy2019md}. On scalability and throughput FeGAN performed better for distributed systems with greater than 32 devices. A %downside of FeGAN is that the entire G and D model weights must be passed between the server and distributed devices. This results in spikes in %bandwidth usage if the model weights are significant.

\subsection{"Synthetic Learning: Learn From Distributed Asynchronized Discriminator GAN
Without Sharing Medical Image Data"}
\cite{asyndgan} demonstrated a distributed GAN with a central generator and distributed discriminators, similar to\cite{hardy2019md}, but with a more complex architecture. Not only is each generated image identified as real or fake, but additionally \cite{asyndgan} describes how "[t]he discriminator individually quantifies the fake or real value of different small patches in the image." The approach is termed Distributed Asynchronized Discriminator GAN (AsynDGAN). In \cite{fegan}, AsynDGAN's architecture was also used to generate medical images used as the sole training data for a segmentation model making this paper the first to validate GAN-generated data for its utility in medical research.

%\cite{asyndgan}'s motivation is similar to our project's: 1) medical %images are private data protected by HIPPA and IRB processes; 2) while %ethically important, data protections limit access to data leading to %slower research and reproducibility concerns; and 3) it is critical for %generated datasets to capture diverse system dynamics of interest if they %are to be of utility.

\cite{asyndgan} proposed a GAN framework to mimic a multi-hospital system with each hospital providing a discriminator with access to its own unique true images of differing quantities. As \cite{asyndgan} states, "data from different [discriminator worker] nodes are often dissimilar." The authors propose that AsynDGAN provides a method to unify the different datasets by the generator learning from multiple, likely disjoint, discriminators. An interesting finding of \cite{asyndgan} is that an individual discriminator was only able to learn a single mode Gaussian due to its local information limitation, but with distributed discriminators, AsynDGAN was able to capture global information and multiple modes.

%AsynDGAN is a conditional GAN, meaning that the architecture learns multiple labels or masks. An %example of a label or mask in the medical image context could be the location of a tumor in a %brain tumor image, or an organ, e.g. breast, liver, or kidney, for a task to generate images of %nuclei segmentation. AsynDGAN modeled its generators from the ResNet~\cite{he2016deep} %architecture, and the discriminators used the structure of the Patch-GAN~\cite{isola2017image} %discriminators. The Adam optimizer was used. Cross entropy loss was implemented, but %\cite{asyndgan} stressed that other GAN losses could be implemented; for example, Wasserstein %distance or mean squared error.

\cite{asyndgan}'s evaluation of its generated images was focused on how well the generated images performed as training data for segmentation models of medical images, specifically, brain tumor and nuclei segmentation. AsynDGAN was compared on Dice score, Sensitivity, Specificity and 95\% quantile of Hausdorff distance (HD95). Dice and HD95 provide two different measures of the distance between two subsets. A higher Dice score is better, while a lower HD95 score is better. Real brain tumor images and AsynDGAN-generated brain tumor images were evaluated in segmentation experiments. Compared to AsynDGAN-generated images, real brain tumor images had higher Dice scores (0.7485 vs 0.7043) and lower HD95 values (12.85 vs 14.94), but AsynDGAN-images had better scores than a non-distributed GAN, or any of the individual GANs trained from a single discriminator in the distributed architecture. This indicates that AsynDGAN improves upon a single discriminator architecture and that the distributed discriminators with disjoint real data provide valuable information to the distributed GAN.

%For nuclei segmentation, Dice and Aggregated Jaccard Index (AJI) were %compared. Similarly to Dice, AJI gauges the similarity or dissimilarity of %sample sets and is routinely used to compare performance of nuclei %segmentation; higher scores for both are better. Comparing results of a %segmentation model trained on real images vs. AsynDGAN-generated images, %AsynDGAN had a higher Dice score compared to real images (0.7930 vs. %0.7833) and an equal AJI score (0.5608) for nuclei segmentation.

\section{Method}
\label{section:proposal}

\subsection{Identification of novel dataset}
\label{section:dataset}
In order for GAN-generated synthetic data to be a viable open source of healthcare data, the underlying studies must be reproducible. In this vein, we reviewed publicly available datasets on which to train our GAN. For evaluation of the GAN synthetic data, the study must have a sufficiently large sample size so that the study can be reproduced on a subset of the total data in order for a validation set to be held out.

We propose leveraging existing knowledge of tabular GANs, conditional tabular GANs, and distributed GANs to produce a first step towards a generative model capable of producing statistically accurate, but variable tabular data from a small initial dataset using distributed training. The demo eICU\cite{eICU} database provides 2,500 patient-unit stay records and includes patient demographics such as age, ethnicity, and gender, past medical history, medications, initial diagnosis, treatments, and discharge diagnosis. A unit stay is the unique combination of a patient, the specific hospital ICU unit in which they stayed (e.g. Cardiac ICU, Neuro ICU, and Med-Surg ICU), and the start time for the patient in the ICU unit.

In developing discGAN and comparing to the state-of-the-art (SOTA) CTGAN, we evaluated the following features:
\begin{enumerate}
    \item age: the age of the patient when they began their unit stay
    \item hospital discharge offset: the number of minutes from unit admit time to when the patient was discharged from the hospital
    \item CHF: indicator that the patient had a past history of congestive heart failure (CHF)
    \item COPD\_nolimitations, COPD\_moderate, COPD\_severe: indicator that the patient had a past history of mild, moderate, or severe chronic obstructive pulmonary disease (COPD)
    \item NoHealthProblems: indicator that the patient had no past history of health problems
    \item asthma: indicator that the patient had a past history of asthma
    \item homeoxygen: indicator that the patient had a past history of requiring oxygen treatment at home
    \item hypertensionrequiringtreatment: indicator the patient had a past history  hypertension (high blood pressure) requiring treatment
    \item restrictivepulmonarydisease: indicator the patient had a past history of restrictive pulmonary disease
    \item All [past history count]: count of all past history diagnosis entered into the eICU system when the patient was admitted to the hospital unit. These include past histories not listed above.
    \item ethnicity: ethnicity of the patient. Ethnicity categories are: African American, Asian, Caucasian, Hispanic, Native American, and Other/Unknown
    \item gender: gender of the patient. Gender categories are: Male, and Female.
    \item discharge status: whether the patient was alive, or expired when they were released from the hospital unit. Discharge status categories are: Alive, Expired, and Other.
\end{enumerate}

\subsection{Model architecture review}
\label{section:architecture}

%The size of the data (2,500 patient ICU stays) is not adequate for robust conditional data generation; however we %will apply two techniques to address the paucity of data. Instead of generation for only the raw data as input, we %will sample from the conditioned kernel density estimation (KDE) and use the samples as generator input. 

% Include this second approach if we have time
%A second approach, is to create a Variational Autoencoder (VAE) and generate a %latent space for the raw data, then use the latent space as input to our GAN. The %GAN output can then be decoded, creating our synthetic healthcare data.

We will use methods described in \cite{xu2019modeling} to produce synthetic, single-table synthetic data. In particular, CTGAN overcomes several issues related to using GANs to generate synthetic tabular data; namely, "the need to simultaneously model discrete and continuous columns, the multi-modal non-Gaussian values within each continuous column, and the severe imbalance of categorical columns." \cite{xu2019modeling} In contrast to "vanilla" GANs, CTGAN uses mode-specific normalization to overcome the issue of multi-modal, non-Gaussian distributions, and address data imbalance with a conditional generator and training-by-sampling. The overall architecture implements "fully-connected networks and several recent techniques to train a high-quality model." \cite{xu2019modeling} Additionally, CTGAN addresses the issue of class-imbalanced data when training the generator, which is unusual in a typical GAN architecture. CTGAN aims to "resample efficiently in a way that all the categories from discrete attributes are sampled evenly [during training] [...] and to recover the (not sampled) real data distribution during testing."\cite{xu2019modeling} Thus, the data generation is conditional on each particular column value. 

\subsection{discGAN Approach}
\label{section:approach}
We generate tabular healthcare data using a  distributed, conditional GAN, an appropriate tool to address our motivation because, in its simplest implementation, a GAN directs a zero-sum game in which one model, the generator, produces synthetic data and another, the discriminator, determines whether the generated data are true or erroneous. This process happens iteratively, with both models training based on the outcome of each game. In our selected domain, a GAN must be able to generate data for selected conditions; for example, it must handle not only the patient age distribution for the entire hospital, but the patient age distribution specific to the Cardiac ICU. This requires a GAN that can produce conditioned synthetic data. A conditional tabular GAN is trained similarly to a basic tabular GAN, with the additional necessity of learning the data distributions by class. 
%The differences in training are shown in Figure \ref{fig:trainingGAN}.

We call our implementation discGAN for \emph{distributed conditional GAN}.

\subsubsection{Data Pre-Processing}

discGAN expects a \texttt{TensorFlow} DataSet object. As such, our code required minimally pre-processed eICU data. We imported the eICU data via a CSV file into a \texttt{pandas} data frame, standardized the continuous variable, and one-hot encoded all discrete variables using \texttt{sklearn.preprocessing}, converting each feature to a \texttt{NumPy} array with each transformation. Lastly, we concatenated the scaled continuous arrays and one-hot encoded discrete arrays into a \texttt{TensorFlow} DataSet object.  

\subsubsection{discGAN Architecture}

discGAN is relatively lightweight. In particular, we forewent adding a residual block(s) to our model, which is opposed to the SOTA CTGAN implementation, as our model performed well without one. Both discriminator and generator utilize a standard Adam optimizer to update the model weights. 

The discGAN generator contains an input layer followed by three sets of: (1) a dense layer, (2) leaky ReLU activation, and (3) batch normalization. After this series, a final leaky ReLU activation is incorporated before a final dense layer to create the output. Overall, there are 18,945 trainable parameters and 384 non-trainable parameters in the generator. 

The discGAN discriminator has an even simpler architecture. Its input layer is followed by two triplets of (1) a dense layer, (2) leaky ReLU activation, and (3) dropout regularization before a final dense output layer. Overall, there are 7,105 trainable parameters and 0 non-trainable parameters in the discriminator. 

Our model is malleable in its distributing capabilities. It is built so that it can be run without distributing, with either the generator or discriminator distributed, or with both distributed. We successfully provisioned multiple GPUs as multiple workers. In a true distributed environment, we imagined that dozens or hundreds of workers may be provisioned and we expect our methods to scale as described in \cite{fegan}, \cite{hardy2019md}, and \cite{asyndgan}.
discGAN's implementation of distribution and our distribution of the SOTA CTGAN is describe in section \ref{distribute}.

%\begin{figure}
%     \centering
%     \begin{subfigure}[b]{0.45\textwidth}
%         \centering
%         \includegraphics[width=\textwidth]{images/basic_gan (2).png}
%         \caption{Basic GAN}
%         \label{fig:basicGAN}
%     \end{subfigure}
%     %\hfill
%     \begin{subfigure}[b]{0.45\textwidth}
%         \centering
%         \includegraphics[width=\textwidth]{images/conditional_gan (1).png}
%         \caption{Conditional GAN}
%         \label{fig:conditionalGAN}
%     \end{subfigure}
%     
%     \caption{Training tabular GANs}
%    \label{fig:trainingGAN}
%\end{figure}

\subsection{Evaluating generated tabular data}
\label{eval_description}

\paragraph{Qualitative evaluation} Initial model development of 1D and 2D GANs utilized qualitative and simple quantitative evaluation of generated data. Intra-training distribution plots as seen in Figure \ref{fig:trainingAgeGAN} were used for evaluation. These plots demonstrate the GAN learning the data distribution(s); the generated data becomes visually similar to the true distribution throughout training. In addition to subjective comparison of histograms, the generated minimum, maximum, mean, and standard deviations of real and generated data were compared.

\paragraph{Quantitative evaluation} To compare 2D and $n$-dimensional tabular data, we compared distribution of continuous and discrete columns from synthetic and real data. For continuous columns we use the two-sample Kolmogorov-Smirov (KS) test \cite{kstest}. The two samples are the real and the synthetic data. The KS test value is 1 minus the KS Test D statistic, which indicates the maximum distance between the real cumulative distribution function (CDF) and the synthetic CDF.\cite{patki2016synthetic} The average KS Test value for all continuous columns in the real vs. generated data is used  as a statistical metric of the performance of the GAN on continuous columns. 

For discrete columns, the chi-squared test compares the real and generated discrete distributions. The p-value of the chi-squared test is used to indicate the probability that the two discrete columns are from the same distribution.\cite{patki2016synthetic} The mean probability for discrete columns is returned as the CS Test metric.

Additionally, we compare the ability of the generated vs. real data to be used in machine learning algorithms. \cite{patki2016synthetic}'s \emph{machine learning efficacy}\cite{patki2016synthetic, tabGANreview} metrics indicate if the generated data is comparably useful as training data for a machine learning task. We compare the performance of a decision tree classifier (Decision Tree Classifier) and a multilayer perceptron classifier (MLP Classifier) to classify a target column. Generated data's \emph{machine learning efficacy} measures are compared to the same measure for a train/test split of real data.

In section \ref{evalandcompare} KS Test, CS Test, Decision Tree Classifier and MLP Classifier are evaluated for:
\begin{itemize}
    \item an 80/20 train/test split of real data
    \item real vs. non-distributed, and distributed discGAN generated data
    \item real vs. non-distributed and distributed CTGAN generated data
\end{itemize}

We propose three comparison metrics to evaluate the difference in KS Test, CS Test and machine learning efficacy between real vs. synthetic data, and real vs. a train/test split of real data. KS Test Compare, KSTC (equation \ref{kstc}), is used to evaluate the distance between the synthetic and real data, vs. the distance between a train/test split of the real data. CS Test Compare, CSTC (equation \ref{cstc}), is used to compare the probability of the synthetic and real data belongs to the same distribution vs. train/test split data belonging to the same distribution. Finally, the machine learning efficacy comparison, MLEC (equation \ref{mlec}), is used to evaluate the relative performance of generated data for an ML classification task, to that of a train/test split of real data.

% \begin{multicols}{2}

\begin{equation} \label{kstc}
	\textrm{KSTC} = 1 - \bigg | \bigg ( 1 - \frac{\textrm{KSTest}_{generated}}{\textrm{KSTest}_{real}} \bigg )\bigg |
\end{equation}

\begin{equation} \label{cstc}
	\textrm{CSTC} = 1 - \bigg | \bigg ( 1 - \frac{\textrm{CSTest}_{generated}}{\textrm{CSTest}_{real}} \bigg )\bigg |
\end{equation}

% \columnbreak

\begin{equation} \label{mlec}
	\textrm{MLEC} = 1 - \bigg | \bigg ( 1 - \frac{\textrm{MLE}_{generated}}{\textrm{MLE}_{real}} \bigg )\bigg |
\end{equation}

% \end{multicols}

In Section \ref{evalandcompare}, equations \ref{kstc}, \ref{cstc}, and \ref{mlec} are used to evaluate both discGAN and CTGAN synthetic records.

\section{Preliminary Experiments}

\subsection{Proof of concept: a one-dimensional GAN}
\begin{figure}[H]
\centering
%\begin{adjustwidth}{50em}{0em}
  \begin{subfigure}[b]{0.23\textwidth}
    \includegraphics[width=\textwidth]{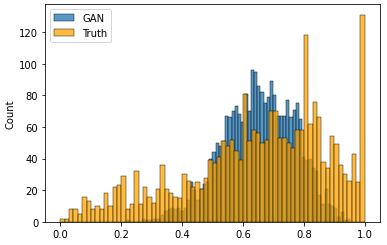}
    \caption{1 training iteration}
  \end{subfigure}
  \begin{subfigure}[b]{0.23\textwidth}
    \includegraphics[width=\textwidth]{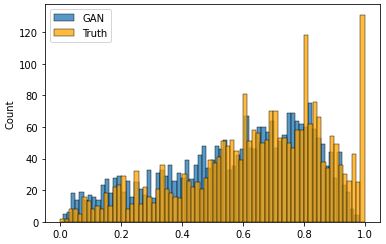}
    \caption{5,000 training iterations}
  \end{subfigure}
%\end{adjustwidth}
%\begin{adjustwidth}{-1em}{0em}
  \begin{subfigure}[b]{0.23\textwidth}
    \includegraphics[width=\textwidth]{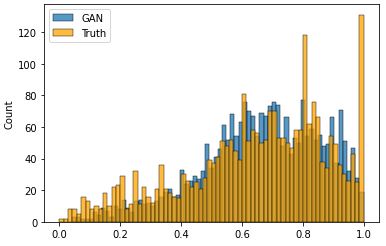}
    \caption{10,000 training iterations}
  \end{subfigure}
  \begin{subfigure}[b]{0.23\textwidth}
    \includegraphics[width=\textwidth]{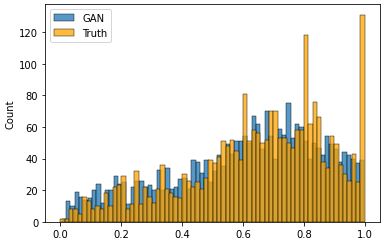}
    \caption{15,000 training iterations}
  \end{subfigure}
%\end{adjustwidth}

\caption{Real (orange) and generated (blue) age distributions while training the 1D age GAN.}
\label{fig:trainingAgeGAN}
\end{figure}

For a first proof of concept, we produced a non-conditional GAN to generate a one-dimensional dataset of eICU patient ages. The eICU dataset contains 2,500 discrete patient ages from 15 - 90, with a mean of 63.3 and standard deviation of 17.72. The original dataset combined all patient ages greater than 90 into a single category `90+'; these records were given the age of '90'. The true age distribution can be seen in orange in Figure \ref{fig:trainingAgeGAN}.

A basic GAN is used to generate the one-dimensional age distribution. The discriminator input is a single value for a generated age, followed by 1) a 64-node dense layer with LeakyReLU activation (alpha=0.2), 2) a 10\% dropout layer, 3) another 64-node dense layer with LeakyReLU activation (alpha=0.2), 4) a 10\% dropout layer, and 5) a 1-node output layer with sigmoid activation indicating the probability of the input being real.

The generator has a random noise input of dimension 50 followed by 1) a 64-node dense layer with LeakyReLU activation (alpha=0.2), 2) a 64-node dense layer with LeakyReLU activation (alpha=0.2), and 3) a 1-node output layer with sigmoid activation. Patient age was normalized, thus, the output layer with sigmoid activation produced a synthetic age after inverse scaling.

After 15,000 iterations the results of the 1D Gan were acceptable by visual inspection of the resulting histograms of synthetically generated ages vs. real ages. The in-progress training and comparison of real and GAN-generated ages are shown in Figure \ref{fig:trainingAgeGAN}. The mean of the generated ages was 60.2 years (compared to the real data mean of 63.3) and the standard deviation of the generated ages was 17.99 (compared to real data standard deviation of 17.72).

\subsection{Proof of concept: a conditional tabular GAN}

%A conditional GAN's generator must be able to generate tabular data, both %continuous and discrete, from multiple classes, and the GAN's %discriminator must be able to correctly classify real and fake data from %multiple classes. Our GANs normalize data, and one-hot encode categories %in both the generator and discriminator. 

As our second proof of concept, we implemented a 2D conditional GAN architecture for generating 1) age-by-ethnicity for Caucasian, African American and Native American ethnicity, and 2) for age-by-unit-type for CTICU, CSICS, and Cardiac ICU unit types. Samples in our real data were significantly skewed by condition count, and had different means and standard deviations as shown in Tables \ref{age_condition_unit} and \ref{age_condition_ethnicity}. These results are encouraging as the generated data both quantitatively and qualitatively resembles real data; however, it is noteworthy that the standard deviation for all unit types and ethnicities is smaller in the generated data than real data, significantly so for the Native American class, which had only 12 real examples in our training data.

% \begin{multicols}{2}

\begin{table}
  \caption{Real and Generated Age by ICU Unit Type}
  \label{age_condition_unit}
  \centering
  \begin{tabular}{llllll}
    \toprule
     &
    \multicolumn{3}{r}{Real\,\,\,\,\,\,\,\,\,\,\,} & \multicolumn{2}{c}{Generated}                   \\
    \cmidrule(r){1-6}
    Unit Type     & Count     & Mean    & Stdv &  Mean & Stdv \\
    \midrule
    Cardiac ICU & 133  & 60.2 & 17.1 & 58.9 & 16.7 \\
    CTICU ICU     & 52       & 61.7 & 10.3 & 61.7 & 9.4 \\
    CSICU     & 65 & 67.6 & 9.3 & 68.3 & 9.3  \\
    \bottomrule
  \end{tabular}
  % \vspace{-4mm}%Put here to reduce too much white space after your table 
\end{table}

\begin{table}
  \caption{Real and Generated Age by Ethnicity}
  \label{age_condition_ethnicity}
  \centering
  \begin{tabular}{llllll}
    \toprule
     &
    \multicolumn{3}{r}{Real\,\,\,\,\,\,\,\,\,\,\,} & \multicolumn{2}{c}{Generated}                   \\
    \cmidrule(r){1-6}
    Ethnicity     & Count     & Mean    & Stdv &  Mean & Stdv \\
    \midrule
    African American & 230  & 56.2 & 16.8 & 54.4 & 15.9 \\
    Caucasian     & 2010 & 64.4 & 17.4    & 64.4 & 16.1 \\
    Native American     & 12       & 50.5 & 19.5  & 50.0 & 9.8 \\
    \bottomrule
      \end{tabular}
  % \vspace{-4mm}%Put here to reduce too much white space after your table 
\end{table}

% \end{multicols}

A 2D discGAN uses a default noise input of 50 followed by 2 64-node dense layers with LeakyReLU activation and a dropout layer (dropout rate of 0.1). Batch sizes of 16, 32, and 64 were used, and 32 produced the most accurate results. 

The following figures contain the output of the 2D GAN:

%\begin{figure}[H]
%    \centering
%    \includegraphics[width=16cm]{unit_age.png}
%    \caption{discGAN generated age and ICU unit type data}
%    \label{fig:age_ethnicity}
%\end{figure}

% \begin{figure}[H]
%     \centering
%     \includegraphics[width=14cm]{ethnicity_age.png}
%     \caption{discGAN generated age and ethnicity data}
%     \label{fig:age_unit}
%     \vspace{-4mm}
% \end{figure}

\begin{figure}
\centering
%\begin{adjustwidth}{50em}{0em}
  \begin{subfigure}{0.5\textwidth}
    \includegraphics[width=\textwidth]{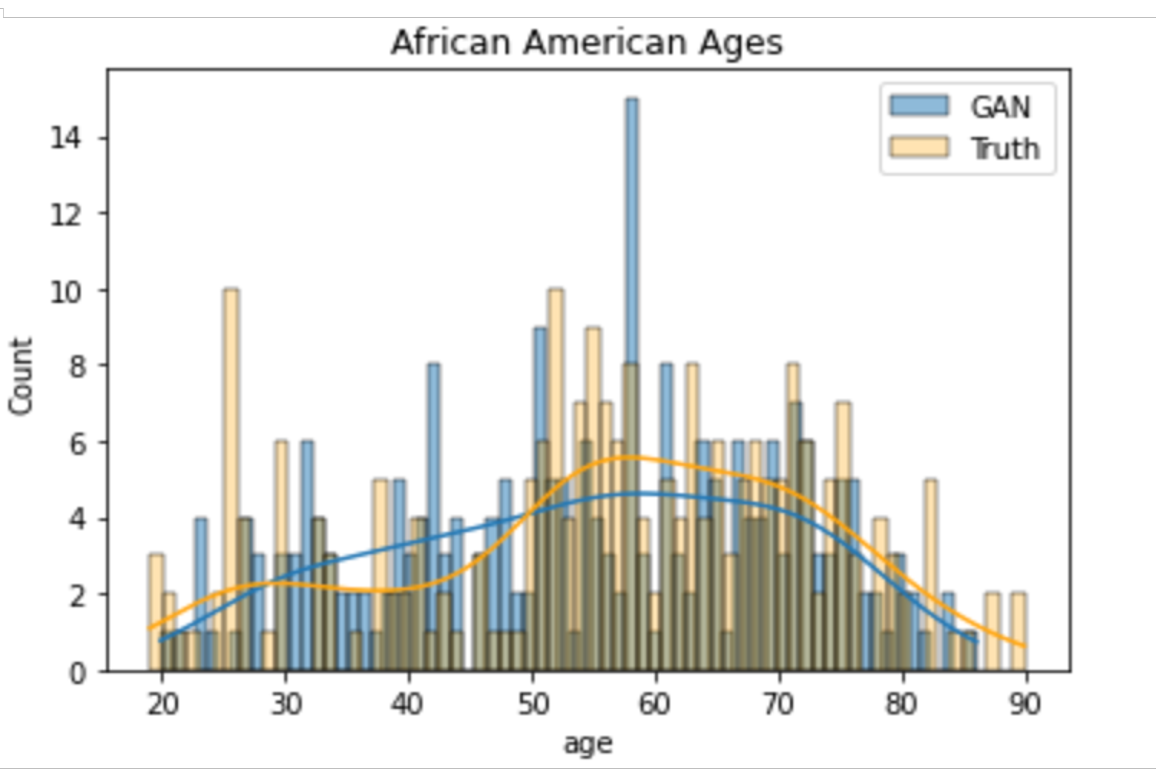}
    % \caption{1 training iteration}
    \label{fig:age_unit_1}
  \end{subfigure}
  \begin{subfigure}[b]{0.5\textwidth}
    \includegraphics[width=\textwidth]{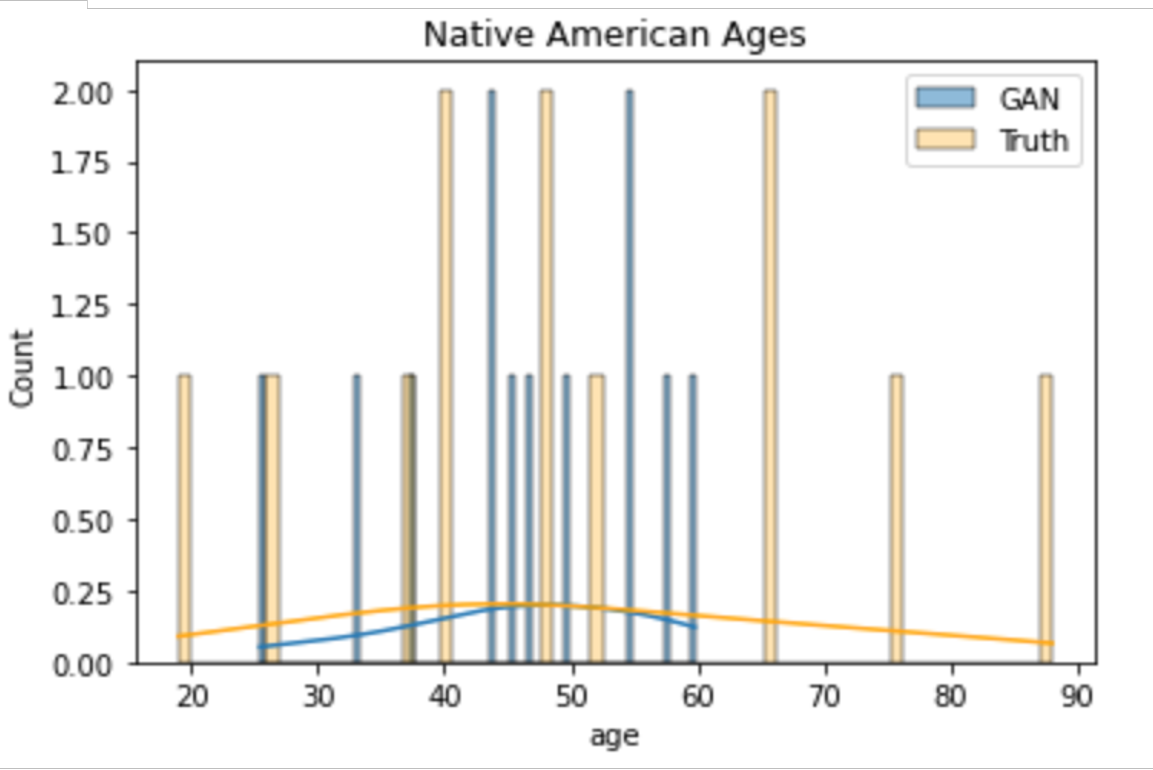}
    % \caption{5,000 training iterations}
    \label{fig:age_unit_2}
  \end{subfigure}
%\end{adjustwidth}
%\begin{adjustwidth}{-1em}{0em}
  \begin{subfigure}[b]{0.5\textwidth}
    \includegraphics[width=\textwidth]{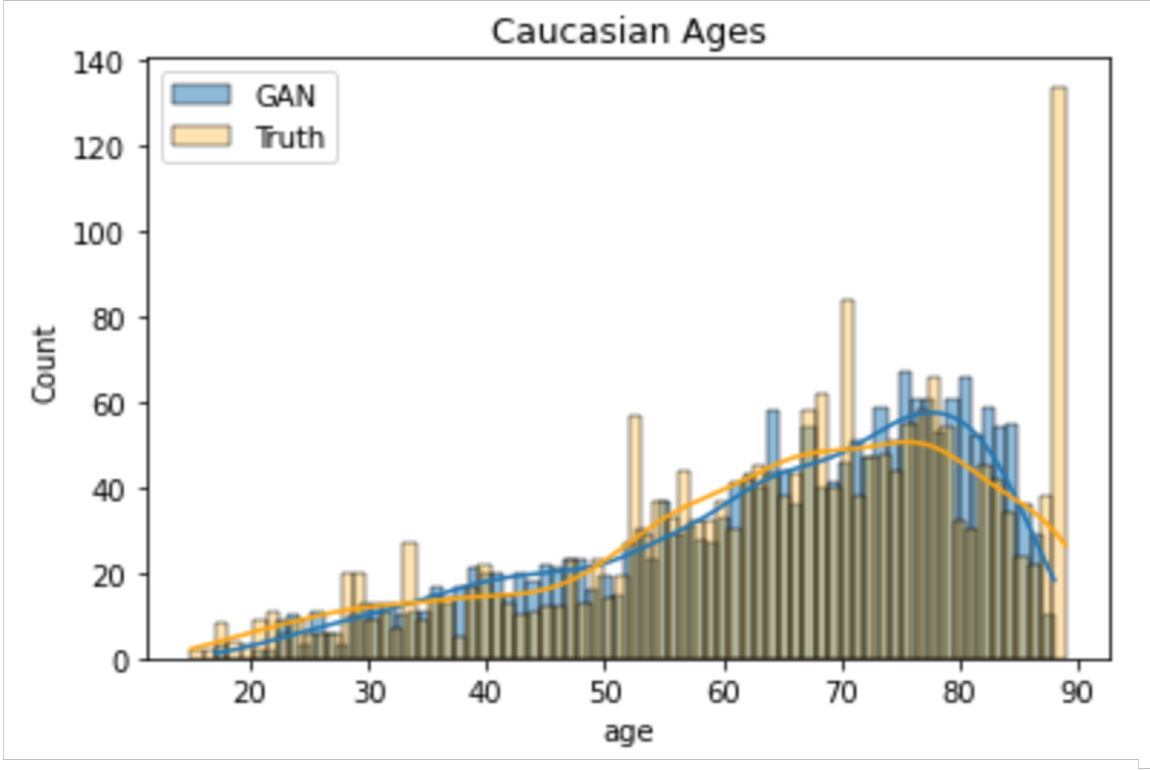}
    % \caption{10,000 training iterations}
    \label{fig:age_unit_3}
  \end{subfigure}
\caption{discGAN generated age and ethnicity data}
\label{fig:age_unit}
\end{figure}

\label{distribute}
\subsection{Distributing a tabular GAN}
discGAN models are written with the \texttt{TensorFlow} library \cite{tensorflow2015-whitepaper}, which provides the tf.distrbute\cite{tensorflow} API to distribute model training across GPUs and machines. To demonstrate a tabular GAN in a distributed environment, we provisioned two GPU's and utilized tf.distribute's \texttt{MultiWorkerMirroredStrategy} which enables synchronous and distributed training using multiple GPUs on a single machine. \texttt{MultiWorkerMirroredStrategy} can distribute training across multiple workers, each with "potentially multiple GPUs."\cite{tensorflow} More specifically, the strategy replicates and updates all model parameters, weights, and computations to each device (hence it being a "mirrored" strategy), trains on each device, then uses distributed collective implementation -- namely \emph{all-reduce}.  \emph{All-reduce} is a fused algorithm that \texttt{TensorFlow} uses to "aggregate tensors across all the devices by adding them up, and makes them available on each device."\cite{tensorflow} \texttt{TensorFlow} uses the NVIDIA Collective Communications Library implementation of the all-reduce algorithm.

The strategy distributes discGAN by allowing the GPU workers to train together. Specifically, the devices divide the training data at each step before running forward and backward propagation separately on a mirrored copy of the the GAN models and their parameters on each device. The strategy then uses all-reduce to aggregate the distributed gradients and sends updated variable weights and a single aggregated gradient to all devices before starting the next training step.

As mentioned in \ref{section:approach}, we built discGAN to distribute the generator or discriminator individually, or both together. We focused on distributing the discriminator in our experiments, with results described in \ref{evalandcompare}.

Additionally, for comparative purposes, we successfully distributed CTGAN. Note that we utilized the \texttt{nn.DataParallel} class to distribute CTGAN since it is written in PyTorch rather than \texttt{TensorFlow}. \texttt{nn.DataParallel} follows a similar distribution implementation where models are trained on batches of training data across machines. During each training iteration, a "main" GPU broadcasts updated model parameters to each worker GPU before forward propagation is run on mini-batches of training data. Each output is sent to the main GPU, where the network loss is calculated and sent to the worker GPUs, which then run back prop. Finally, the main GPU runs all-reduce to average gradients before the updated model is sent to each worker GPU prior to the next training iteration. \cite{NEURIPS2019_9015, mohan_2019}

\section {Results}
\label{evalandcompare}

\subsection{discGAN results}

With discGAN we generated multiple discrete columns in our tabular data with a single continuous column. We selected hospital discharge offset, which is the number of minutes from unit admit time that the patient was discharged from the hospital. The discrete variables included in the real data were CHF, COPD\_moderate, COPD\_nolimitations, COPD\_severe, NoHealthProblems, asthma, homeoxygen, hypertensionrequiringtreatment, restrictivepulmonarydisease, All (count of all past history entries for the patient), ethnicity, gender, and dischargestatus. discGAN is unable to generate representative data with multiple continuous features. Hospital discharge offset was selected as our continuous feature, thus the KS Test value is for hospital discharge offset in the generated data vs. real data. CS Test value is the mean chi-squared test p-value for all the discrete columns in the generated data vs. real data. Machine learning efficacy, CS Test, and KS Test values were evaluated from 249,000 generated samples from the trained discGAN model. In comparison, the real eICU dataset contains 2027 records. 249,000 synthetic records were generated in order to verify discGAN can generate substantial number of statistically similar synthetic data records from a small dataset.

\begin{figure}
\centering
%\begin{adjustwidth}{50em}{0em}
  \begin{subfigure}[b]{0.48\textwidth}
    \includegraphics[width=\textwidth]{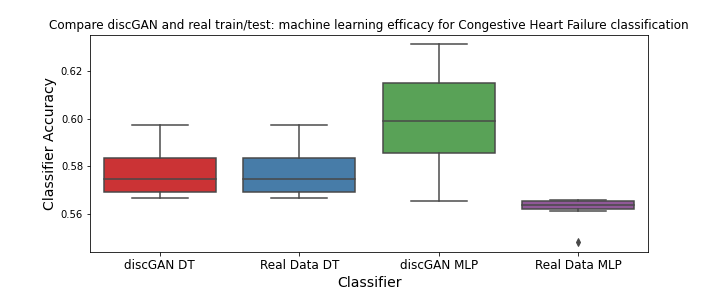}
    \caption{ML Efficacy for CHF classification}
  \end{subfigure}
  %
%\begin{adjustwidth}{-1em}{0em}
  \begin{subfigure}[b]{0.48\textwidth}
    \includegraphics[width=\textwidth]{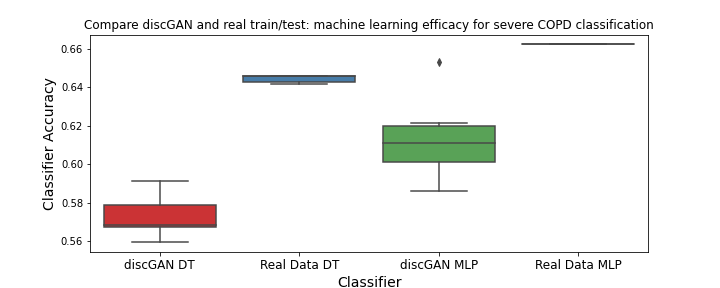}
    \caption{ML Efficacy for COPD\_severe classification}
  \end{subfigure}
%\end{adjustwidth}

\caption{discGAN machine learning efficacy: (a) CHF machine learning efficacy: discGAN  (red) and real (blue) data classification accuracy of Decision Tree classifiers for single-GPU and two-GPU discGANs. Results include epochs of 15,000, 20,000 and 25,000. discGAN (green) and real (purple) data classification accuracy of MLP classifiers for single-GPU and two-GPU discGANs. Results include epochs of 15,000, 20,000 and 25,000. \\
(b) Severe COPD machine learning efficacy:  discGAN  (red) and real (blue) data classification accuracy of Decision Tree classifiers for single-GPU and two-GPU discGANs. Results include epochs of 15,000, 20,000 and 25,000. discGAN (green) and real (purple) data classification accuracy of MLP classifiers for single-GPU and two-GPU discGANs. Results include epochs of 15,000, 20,000 and 25,000.} 
\label{fig:discGAN_classifiers}
\vspace{-10mm}%Put here to reduce too much white space after your table
\end{figure}

Figure \ref{fig:discGAN_classifiers} shows the results of machine learning efficacy tests for past histories of congestive heart failure and severe COPD. Figures (a) and (b) show comparisons of discGAN generated data used as training data, and real data used as test data, vs an 80/20 train/test split of real data. Figure \ref{fig:classifiers} (a) shows that discGAN was able to generate joint distributions of all features with similar machine learning efficacy for congestive heart failure using a decision tree classifier. The generated data most closely matched the decision tree classifier efficacy; however, neither the real nor generated data were effective for classifying congestive heart failure.

Figure \ref{fig:discGAN_classifiers} (b) shows the results of the machine learning efficacy tests for the classification of a past history of severe COPD. The efficacy of the discGAN generated data vs. real data for severe COPD classification are less similar than for congestive heart failure, and interestingly, the discGAN generated data appears relatively more effective for classifying severe COPD by an MLP classifier. While more effective, this indicates the discGAN-generated data may not be statistically similar to the real data.

\subsection{CTGAN results}
With CTGAN we generated age, hospital discharge offset, CHF, COPD\_moderate, COPD\_nolimitations, COPD\_severe, NoHealthProblems, asthma, homeoxygen, hypertensionrequiringtreatment, restrictivepulmonarydisease, All (count of all past history entries for the patient), ethnicity, gender, and dischargestatus. Unlike with discGAN, CTGAN is able to generate multiple continuous features, in our case age and hospital discharge offset, thus the KS Test value is the average KS Test value for age and hospital discharge offset in the generated data vs. real data. Machine learning efficacy, CS Test, and KS Test values were evaluated from 5000 generated samples from the trained CTGAN model. In comparison, the real eICU dataset contains 2027 records. 5000 synthetic records were generated in order to verify that a substantial number of statistically similar synthetic data records could be generated by CTGAN from an initially small dataset.

\begin{figure}
\centering
%\begin{adjustwidth}{50em}{0em}
  \begin{subfigure}[b]{0.48\textwidth}
    \includegraphics[width=\textwidth]{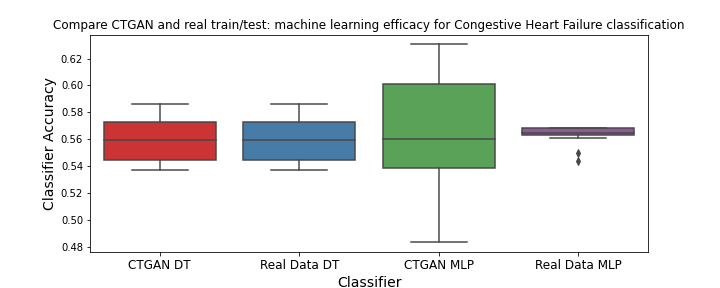}
    \caption{ML Efficacy for CHF classification}
  \end{subfigure}
  %
%\begin{adjustwidth}{-1em}{0em}
  \begin{subfigure}[b]{0.48\textwidth}
    \includegraphics[width=\textwidth]{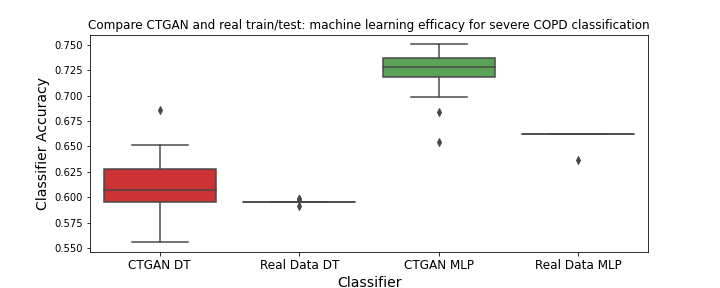}
    \caption{ML Efficacy for COPD\_severe classification}
  \end{subfigure}
%\end{adjustwidth}

\caption{CTGAN machine learning efficacy: (a) CHF machine learning efficacy: CTGAN  (red) and real (blue) data classification accuracy of Decision Tree classifiers for single-GPU and two-GPU CTGANs. Results include epochs of 15,000, 20,000 and 25,000. CTGAN (green) and real (purple) data classification accuracy of MLP classifiers for single-GPU and two-GPU CTGANs. Results include epochs of 15,000, 20,000 and 25,000. \\
(b) Severe COPD machine learning efficacy:  CTGAN  (red) and real (blue) data classification accuracy of Decision Tree classifiers for single-GPU and two-GPU CTGANs. Results include epochs of 15,000, 20,000 and 25,000. CTGAN (green) and real (purple) data classification accuracy of MLP classifiers for single-GPU and two-GPU CTGANs. Results include epochs of 15,000, 20,000 and 25,000.} 
\label{fig:classifiers}
% \vspace{-10mm}%Put here to reduce too much white space after your table
\end{figure}

Figure \ref{fig:classifiers} shows the results of machine learning efficacy tests for past histories of congestive heart failure and severe COPD. Figures (a) and (b) show comparisons of CTGAN generated data used as training data, and real data used as test data, vs an 80/20 train/test split of real data. Figure \ref{fig:classifiers} (a) shows CTGAN was able to generate joint distributions of features with similar machine learning efficacy for CHF. The generated data most closely matched the decision tree classifier efficacy; however, neither the real nor generated data were effective for classifying CHF. (b) shows the results of the machine learning efficacy tests for the classification of a past history of severe COPD. The efficacy of the CTGAN-generated data vs. real data for severe COPD classification are less similar than for CHF, and interestingly, the CTGAN generated data appears relatively more effective for classifying severe COPD by an MLP classifier. While more effective, similarly to discGAN, this indicates the CTGAN-generated data may also not be statistically similar to the real data.

\subsection{Comparison: single GPU vs. two GPUs}
\begin{figure}[h]
\centering
% % \begin{adjustwidth}{4em}{0em}
  \begin{subfigure}[b]{0.45\textwidth}
    \includegraphics[width=\textwidth]{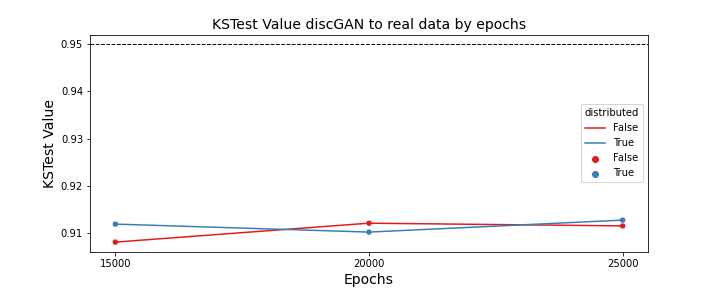}
    \caption{KS Test by epoch: two GPUs (blue), single GPU (red)}

  \end{subfigure}
 %\end{adjustwidth}
  %
%\begin{adjustwidth}{-1em}{0em}
  \begin{subfigure}[b]{0.45\textwidth}
    \includegraphics[width=\textwidth]{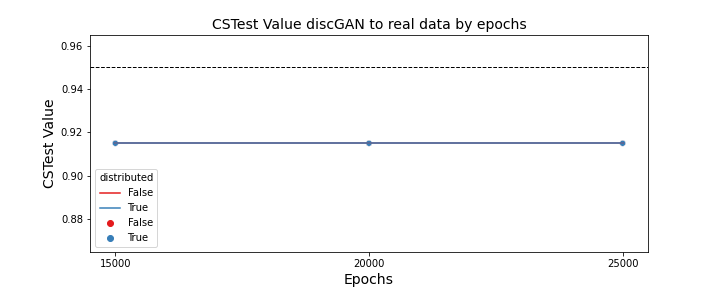}
    \caption{CS Test by epoch: two GPUs (blue), single GPU (red)}

  \end{subfigure}
% \end{adjustwidth}

% \begin{adjustwidth}{4em}{0em}
  \begin{subfigure}[b]{0.45\textwidth}
    \includegraphics[width=\textwidth]{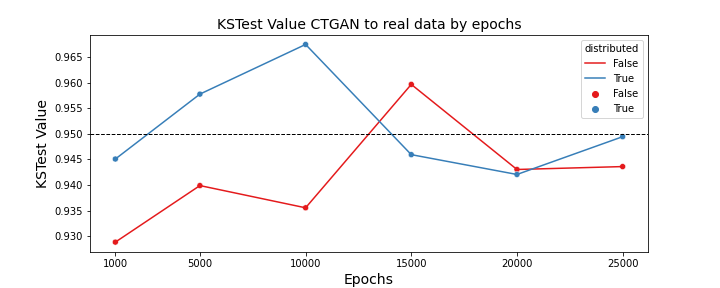}
    \caption{KS Test by epoch: two GPUs (blue), single GPU (red)}

  \end{subfigure}
  %
%\begin{adjustwidth}{-1em}{0em}
  \begin{subfigure}[b]{0.45\textwidth}
    \includegraphics[width=\textwidth]{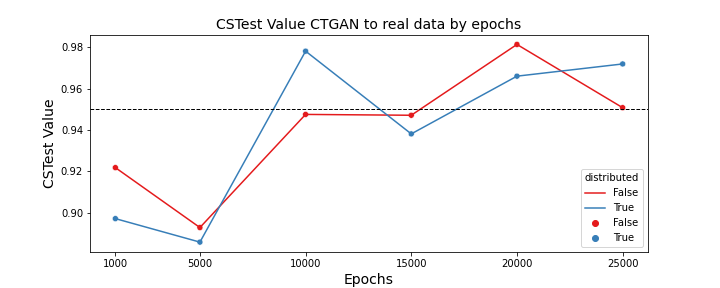}
    \caption{CS Test by epoch: two GPUs (blue), single GPU (red)}

  \end{subfigure}
% \end{adjustwidth}
\caption{(a) and (b) discGAN: KS Test and CS Test values by training epoch for discGAN trained on a single GPU vs. two GPUs evaluated on 249,000 synthetic data rows.\\
\\
(c) and (d) CTGAN KS Test and CS Test values by training epoch for discGAN trained on a single GPU vs. two GPUs evaluated on 5,000 synthetic data rows.} 
\label{fig:GANtrainingSinglev2GPU}
\end{figure}

For both discGAN and CTGAN  we evaluated KS Test and CS Test values by training epochs. Unlike common evaluation tools for image GANs, this comparison provides a quantitative measure of model performance by training time; KS Test and CS Test values were also compared by a GAN trained using a single GPU, and the GAN trained on two GPUs. Figure \ref{fig:GANtrainingSinglev2GPU} displays KS Test and CS Test values for 249,000 discGAN generated synthetic records using 15,000, 20,000, and 25,000 epochs and compares single vs. multi-GPU test values. Figure \ref{fig:GANtrainingSinglev2GPU} displays KS Test and CS Test values for 5,000 CTGAN-generated  synthetic records using 1,000, 5,000, 10,000, 15,000, 20,000, and 25,000 epochs and compares single vs. multi-GPU test values.

% \begin{figure}[h] %{0.45\textwidth}
%     \includegraphics[width=0.5\textwidth]{images/discGAN_KSTest_Value_ours_v_real_249000_byepoch.png}
%     \caption{KS Test by epoch: two GPUs (blue), single GPU (red)}
%     \label{fig:}
% \end{figure}

%\begin{figure}[H]
%\centering
%%\begin{adjustwidth}{50em}{0em}
%  \begin{subfigure}[b]{0.48\textwidth}
%    \includegraphics[width=\textwidth]{images/discGAN_KSTest_Value_ours_v_real_249000_byepoch.png}
%    \caption{KS Test value by epoch: two GPUs (blue), single GPU (red)}
%    \label{fig:}
%  \end{subfigure}
  %
%\begin{adjustwidth}{-1em}{0em}
%  \begin{subfigure}[b]{0.48\textwidth}
%    \includegraphics[width=\textwidth]{images/discGAN_CSTest_Value_ours_v_real_249000_byepoch.png}
%    \caption{CS Test value by epoch: two GPUs (blue), single GPU (red)}
%    \label{fig:}
%  \end{subfigure}
%\end{adjustwidth}

%\caption{discGAN: KS Test and CS Test values by training epoch for discGAN trained on a single GPU vs. two GPUs evaluated on 249,000 synthetic data rows.} 
%\label{fig:discGANtraining}
%\end{figure}

%\begin{figure}[H]
%\centering
%%\begin{adjustwidth}{50em}{0em}
%  \begin{subfigure}[b]{0.48\textwidth}
%    \includegraphics[width=\textwidth]{images/KSTest_Value_ours_v_real_5000_byepoch.png}
%    \caption{KS Test value by epoch: two GPUs (blue), single GPU (red)}
%    \label{fig:}
%  \end{subfigure}
%  %
%%\begin{adjustwidth}{-1em}{0em}
%  \begin{subfigure}[b]{0.48\textwidth}
%    \includegraphics[width=\textwidth]{images/CSTest_Value_ours_v_real_5000_byepoch.png}
%    \caption{CS Test value by epoch: two GPUs (blue), single GPU (red)}
%    \label{fig:}
%  \end{subfigure}
%%\end{adjustwidth}

%\caption{CTGAN: KS Test and CS Test values by training epoch for CTGAN trained on a single GPU vs. two GPUs evaluated on 5000 synthetic data rows.} 
%\label{fig:CTGANtraining}
%\vspace{-4mm}
%\end{figure}

\begin{figure}
    \centering
    \includegraphics[width=0.5\textwidth]{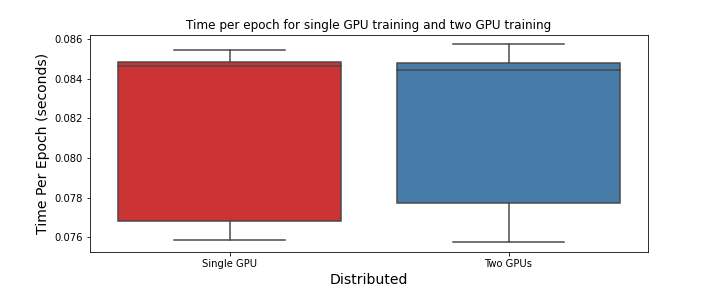}
    \caption{Seconds per epoch training CTGAN on a single GPU and two GPUs.}
    \label{fig:GPUs_training}
\end{figure}

For CTGAN, we evaluated the time-per-epoch training when the GAN was trained on a single GPU and two GPUs. Results are shown in Figure \ref{fig:GPUs_training} and discussed in Section \ref{discussion}.

\subsection{discGAN vs. CTGAN}
\begin{table}[H]
  % \caption{discGAN Evaluation Results}
  \label{table:discGAN_eval}
  \centering
  \begin{tabular}{lcc}
    \toprule
     &    Single GPU & 2-GPU                  \\
    \cmidrule(r){1-3}
    Evaluation Criteria \\
    \midrule
    KS Test & 0.911  & 0.913  \\
    KSTC & 0.922 & 0.923 \\
    CS Test     & 0.915       & 0.915  \\
    CSTC & 0.953 & 0.953 \\
    MLEC Decision Tree Congestive Heart Failure     & 1.0 & 1.0   \\
    MLEC MLP Congestive Heart Failure     & 0.999 & 0.875   \\
    MLEC Decision Tree Severe COPD     & 0.9 & 0.916   \\
    MLEC MLP Tree Severe COPD     & 0.939 & 0.904  \\
    %MLEC Decision Tree Discharge Status     & 0.507 & 0.525   \\
    %MLEC MLP Discharge Status     & 0.532 & 0.564   \\
    %MLEC Decision Tree No Health Problems     & 0.256 & 0.194   \\
    %MLEC MLP No Health Problems     & 0.062 & 0.09   \\
    \bottomrule
  \end{tabular}
  \caption{discGAN evaluation results for 249,000 synthetic records after 25,000 training epochs. Single-GPU and 2-GPU training are compared.}
  % \vspace{-5mm}
\end{table}

\begin{table}[h]% \caption{CTGAN Evaluation Results}
  \label{table:CTGAN_eval}
  \centering
  \begin{tabular}{lcc}
    \toprule
     &    Single GPU & 2-GPU                  \\
    \cmidrule(r){1-3}
    Evaluation Criteria \\
    \midrule
    KS Test & 0.944  & 0.949  \\
    KSTC & 	0.956 & 0.962 \\
    CS Test     & 0.951       & 0.972  \\
    CSTC & 0.99 & 0.988 \\
    MLEC Decision Tree Congestive Heart Failure     & 1.0 & 1.0   \\
    MLEC MLP Congestive Heart Failure     & 0.996 & 0.975   \\
    MLEC Decision Tree Severe COPD     & 0.988 & 0.976   \\
    MLEC MLP Tree Severe COPD     & 0.898 & 0.868   \\
    %MLEC Decision Tree Discharge Status     & 0.559 & 0.555   \\
    %MLEC MLP Discharge Status     & 0.551 & 0.543   \\
    %MLEC Decision Tree No Health Problems     & 0.34 & 0.289  \\
    %MLEC MLP No Health Problems     & 0.09 & 0.046   \\
    \bottomrule
  \end{tabular}
  \caption{CTGAN evaluation results for 5,000 synthetic records after 25,000 training epochs. Single-GPU and 2-GPU training are compared.}
\end{table}

\vspace{10mm}
\label{discussion}
\section{Discussion and Conclusion}
In this paper, we proposed a distributed, conditional tabular GAN, which we call discGAN, to produce synthetic tabular health-care data. Our architecture can be used to handle many discrete features and a single continuous feature, with distribution possible for both the generator and discriminator together or individually. We hope to better generalize our GAN architecture to handle any number of continuous features in future work. 

We performed several experiments using discGAN in which we generated privatized synthetic data, evaluating performance with both qualitative distribution analysis and quantitative methods, including machine learning efficacy, and Kolmogorov-Smirov and Chi-squared tests. For discrete features such as congestive heart failure, discGAN created distributions similar to the real data. For other discrete features such as severe COPD, discGAN generated data was not similar to the real data, possibly because of noise in the real data due to inherent subjectivity in labeling COPD level. For KS Test and CS Test, discGAN performed similarly to the SOTA CTGAN, but CTGAN was able to generate multiple continuous features, unlike discGAN. 

Distribution of discGAN from a single GPU to two GPUs had limited-to-no effect on  training time or our performance measures, including KS Test, CS Test, and machine learning efficacy. The lack of improvement in training time is likely due to the small size of the eICU dataset, resulting in any GPU acceleration of training being canceled out with the time it takes to communicate between the distributed workers. However, the motivation for a distributed, tabular healthcare GAN is not solely to reduce training time; more importantly, its purpose is to allow privacy of real data across multiple workers.

Interestingly, when we updated the CTGAN source code so that we can distribute its training across GPUs, the distributed CTGAN was able to obtain higher KS Test and CS Test values in fewer epochs than a single-GPU CTGAN. Figure \ref{fig:GANtrainingSinglev2GPU} demonstrates that for CTGAN's generation of 5000 synthetic data records, the distributed CTGAN had the highest KS Test value, and it achieved this value in fewer epochs (10,000 epochs) than the highest KS Test value from the CTGAN trained on the single GPU (15,000 epochs). Similarly, the distributed model obtained its highest CS Test value earlier (10,000 epochs) than the single-GPU-trained CTGAN model (20,000 epochs). This demonstrates the effectiveness of distributed training for a GAN.

% \bibliographystyle{ieeetr}
\bibliography{References}

\end{document}